\documentclass[10pt,twocolumn]{article}
\usepackage{cvpr}
\usepackage{times}
\usepackage{epsfig}
\usepackage{graphicx}
\usepackage[cmex10]{amsmath}
\usepackage{amssymb, multirow}
\usepackage{enumitem}

\usepackage[outdir=./]{epstopdf}
\usepackage{algorithm,algorithmic}
\usepackage[caption=false,font=footnotesize]{subfig}
\usepackage{booktabs}
\usepackage{url}

\usepackage{cancel}

\usepackage{chngcntr}
\counterwithin{table}{section}
\counterwithin{figure}{section}

\DeclareGraphicsExtensions{.eps}
\graphicspath{ {images/}, {qualitative/}, {framework/}}

\usepackage[pagebackref=true,breaklinks=true,letterpaper=true,colorlinks,bookmarks=false]{hyperref}

\cvprfinalcopy % *** Uncomment this line for the final submission

\newcommand{\squeezeup}{\vspace{-2.5mm}}

\def\cvprPaperID{272} % *** Enter the CVPR Paper ID here

\ifcvprfinal\fi
\begin{document}

%%%%%%%%% TITLE
\title{Pose2Instance: Harnessing Keypoints for Person Instance Segmentation}

\author{Subarna Tripathi\\
\footnotemark \\
UC San Diego
\\
{\tt\small stripathi@ucsd.edu}
\and
Maxwell D. Collins\\
Google Inc.\\
{\tt\small maxwellcollins@google.com}
\and
Matthew Brown\\
Google Inc.\\
{\tt\small mtbr@google.com}
\and
Serge Belongie\\
Cornell University\\
{\tt\small sjb344@cornell.edu}
}

\maketitle

\renewcommand*{\thefootnote}{\fnsymbol{footnote}}
\footnotetext[1]{Work done during an internship at Google Research.}
\renewcommand*{\thefootnote}{\arabic{footnote}}
\setcounter{footnote}{0}

%%%%%%%%% ABSTRACT
\begin{abstract}
% In this paper, we show how to use object priors to improve instance segmentation.
% We explore the importance of exploiting the domain knowledge for instance level person segmentation.
% The main idea is to harness the prior for the person instance segmentation task using the notion of distance transform of oracle provided human keypoints or estimated keypoints heatmaps.  
Human keypoints are a well-studied representation of people.
% in images.
We explore how to use keypoint models to improve instance-level person segmentation.
The main idea is to harness the notion of a distance transform of oracle provided keypoints or estimated keypoint heatmaps
as a prior for person instance segmentation task within a deep neural network.
For training and evaluation, we consider all those images from COCO where both instance segmentation and human keypoints annotations are available.
We first show how oracle keypoints can boost the performance of existing human segmentation model during inference without any training. 
Next, we propose a framework to directly learn a deep instance segmentation model conditioned on human pose.
Experimental results show that at various Intersection Over Union (IOU) thresholds,
in a constrained environment with oracle keypoints, 
the instance segmentation accuracy achieves $10\%$ to $12\%$ relative improvements
over a strong baseline of oracle bounding boxes.
In a more realistic environment, without the oracle keypoints, the proposed deep person instance segmentation model conditioned on human pose achieves $3.8\%$ to $10.5\%$ relative improvements comparing with its strongest baseline of a
deep network trained only for segmentation.
   
\end{abstract}

%%%%%%%%% BODY TEXT
\section{Introduction}

% \textbf{mdc}: Cursory introduction to what instance segmentation is and the nature of the problem could go here, especially emphasizing why a) instance segmentation is more difficult than semantic segmentation, and b) why people segmentation might be especially interesting. For (b), can talk about difficulty (people tend to appear in images as part of a group, have complex pose) or utility (usually people are the subject or a salient part of the image they’re in). You have this last point in the start, but it is perhaps over-broad, as lots of Computer Vision problems have applications in those same areas.

The instance segmentation problem deals with the pixel-wise delineation of multiple objects, combining segment-level localization and per-pixel object category classification.
This task is more challenging than semantic segmentation, for example the number of object instances is not fixed, unlike the number of object categories. 
Additionally, separating instances that share similar local appearances is highly challenging. 
% TODO(mdc): The hardness of instance segmentation, relative to semantic segmentation, has causes that go somewhat beyond the varying count of instances. For example, consider the problem of distinguishing two adjacent instances of the same class, when different instances can be very similar in local appearance. Compare this to segmenting "car" vs "grass." Perhaps it makes sense to also refer to the current state of the art on these two tasks?
Instance segmentation, in particular person instance segmentation is a promising research frontier for a range of applications such as human-robot interaction, sports performance analysis, and action recognition.  
% Person instance segmentation is even more challenging because in real world, people tend to appear in groups. They exhibit highly complex interaction with other people and objects. Proximity of person instances each exhibiting extremely dynamic poses makes the person instance segmentation more challenging than \emph{generic} object instance segmentation.  

% \begin{figure}
% \begin{center}
% 	\includegraphics[width=1.5in]{keypoints_1.png} 
% % \end{center}
% % \begin{center}
% 	\includegraphics[width=1.5in]{segments_1.png} 
% \end{center}
%    \caption{
% Human keypoints ground truth is easier to collect than precise segmentation masks. Can keypoints help in learning better segmentation? An image from COCO keypoints and segmentation dataset. 
% }
% \label{fig:can_pose_help_segmentation}
% \end{figure}

Deep convolutional neural networks are the current state-of-the-art methods for the task of instance level segmentation. For example, the entrants to the 2016 COCO segmentation challenge \cite{coco_challenge_16_GRMI_seg} achieve excellent performance on instance segmentation for the $80$ object categories considered on the COCO dataset. 
% \textbf{mdc}: Cite “Translation-aware Fully Convolutional Instance Segmentation”and/orMSCOCO2016results. 
Although these methods work extremely well for \emph{any} category of objects, there is a
potential for human-specific domain knowledge to boost the person segmentation performance. 

% In particular, we use the notion of distance transform of oracle provided human keypoints or estimated keypoint heatmaps as a prior for the people instance segmentation task.   
% We call this pose-conditioned segmentation as \emph{Pose2Instance} in this paper. 

\begin{figure*}
\begin{center}
%    \fbox{\includegraphics[width=6.1in]{Pose2Instance_learn.png}}
	\includegraphics[width=6.0in]{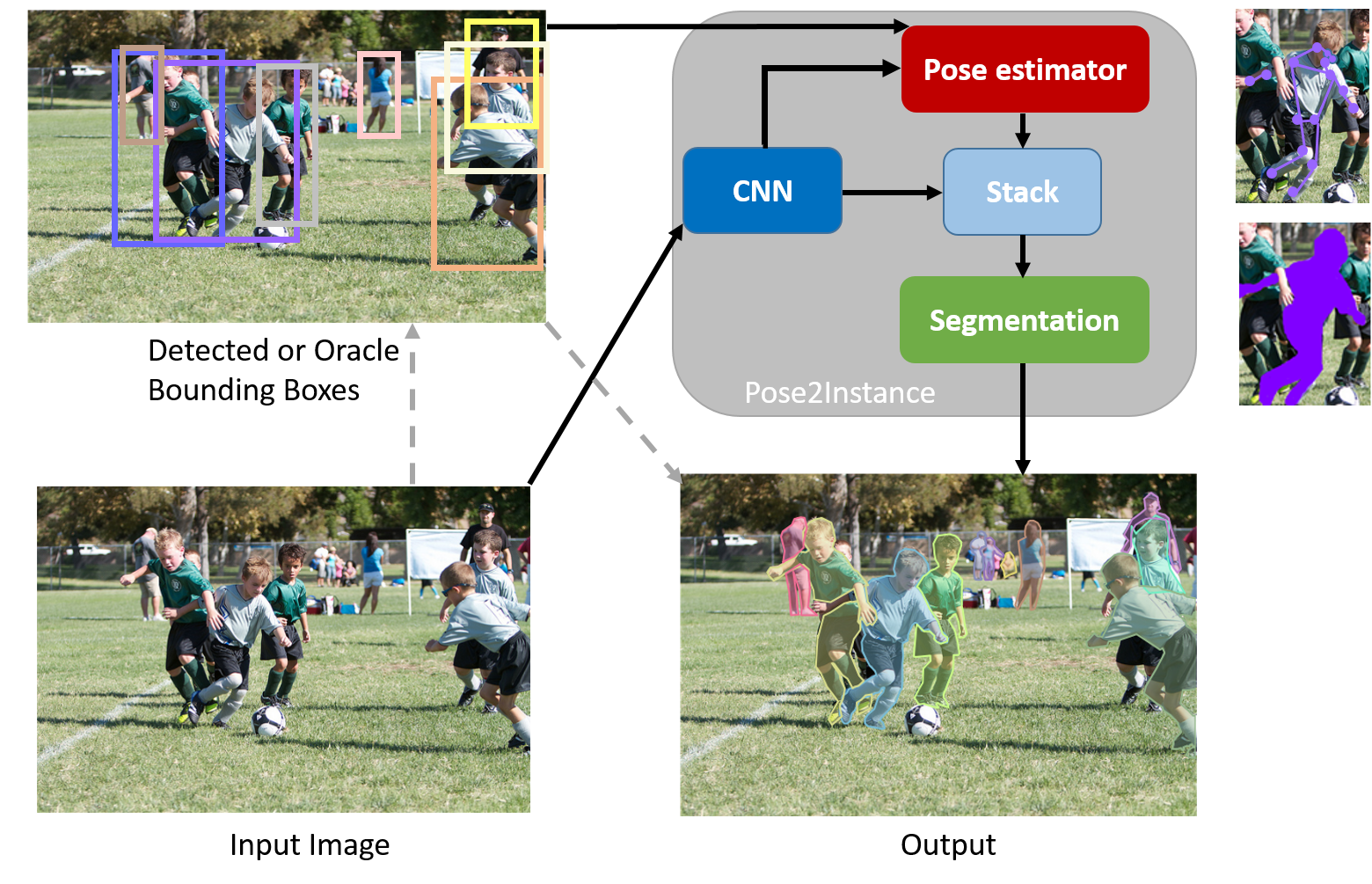}
\end{center}
   \caption{Pose2Instance model for People Instance Segmentation. This model incorporates a \emph{learnable} component conditioned on the human ``pose". The model generates keypoint heatmaps and segmentations at instance level by sharing CNN parameters up to the penultimate layer and uses the keypoints heatmaps output as an  additional input to the segmentation. (best viewed in color)
% An argmax on instance heatmaps of size $h \times w \times n$ generates instance segmentation for the whole image. 
}
\label{fig:Pose2Instance Model1}
\end{figure*}

In this paper, we investigate the importance of human keypoints as a prior for the task of instance-level person segmentation. 
With the availability of image datasets that include both segmentation masks and keypoints annotations, we consider a methodical approach to quantify the importance of keypoints for people instance segmentation. 
We explore what happens if an oracle provides all the keypoints, or only bounding boxes, and how people instance segmentation can be improved respectively. 
% Figure \ref{fig:coco_gt_analysis} shows that human keypoints is an excellent domain knowledge to improve person instance segmentation even atop state-of-the-art deep convolutional neural network based models. 

Our motivation is two-fold. First and foremost, we wish to develop a thorough understanding of whether person-specific domain knowledge is useful for person instance segmentation. 
Second, we wish to quantify the importance of 
human keypoints as a useful domain knowledge for 
improving segmentation over the baseline of best performing deep learning models trained only for segmentation. 
% Third, human keypoints are the most natural representation of a dynamic person model in video. For our futuristic goal of people instance segmentation from such dynamic person models, we want to explore its fundamental stepping stone \ie image-based segmentation conditioned on human pose. 

In order to evaluate the segmentation conditioned on human pose, we consider all instances of people\footnote{We do not include COCO person instances that are marked as "crowd".}
from the COCO segmentation dataset \cite{COCO_eccv14} where the  instances also have keypoints ground truth. By comparing the image and instance identifiers from the COCO segmentation and COCO keypoints dataset, we see there exists $45,174$ images in the training dataset and $21,634$ images in the validation dataset. 
This amounts to $185,316$ and $88,153$ ground truth person instances with both segmentation and keypoints annotations in the training and validation split respectively. 
We call this intersection between COCO instance segmentation dataset and COCO person keypoints dataset as the COCO dataset throughout this paper. 

% \begin{figure*}
% \begin{center}
% % 	\fbox{\includegraphics[width=6.1in]{framework1.png}} 
% 	\includegraphics[width=6in]{framework1.png} 
% \end{center}
%    \caption{Pose2Instance model for learning People Instance Segmentation conditioned on Pose.}
% \label{fig:Pose2Instance Model}
% \end{figure*}

We first explore a human pose prior represented as the distance transform of a skeleton and show how this prior can directly yield instance-level human segmentation when combined with existing semantic segmentation model such as DeepLab \cite{deeplab_chen14semantic} trained for human segmentation. 
This analysis also validates the idea of combining even two existing different models, one for pixel-level person segmentation (non-instance) and another for detecting keypoints, for improving instance segmentation. 

Next, we propose an approach to directly generate the per-pixel probability of people instances conditioned on human poses
% . To learn the segmentation directly conditioned on the estimated keypoints, we propose a framework 
% distance transform of the key poses and pose-conditioned segmentation of individual people 
using a deep convolutional neural network (CNN). 
We call this pose-conditioned segmentation model \emph{Pose2Instance}.
Figure \ref{fig:Pose2Instance Model1} outlines the approach.
Person instance bounding boxes are either provided by an
oracle or they come from a person detector.
The model is trained for generating keypoints heatmaps and segmentation at instance level by sharing \emph{cnn} parameters up to the penultimate layer and using the keypoints heatmap as an additional input channel for the segmentation output. 
% An argmax on instance heatmaps of size $h \times w \times n$ generates instance segmentations for the whole image. Where $h$, $w$ and $n$ denote height, width of the image and the number of instances respectively.

\noindent
In summary, we contribute the following. 
\squeezeup
\begin{itemize} 
% [leftmargin=-.01in]

\item We show that human pose prior represented as the distance transform of the human skeleton yields significant performance gain for the deep people instance segmentation during inference without any training. 
% We show how the oracle keypoints can boost the performance of existing human segmentation model. 

% ST: (1) Deep models are capable of extracting features from big support area => how and whether additional information help?
% (2) The inference should be a single forward pass through the DCNN. Finding a MAP solution like [15] does in a pose-specific CRF can not be extended for CNN, as that becomes computationally prohibitive.

% \item We validate the idea of combining even two existing different models one for pixel-level person segmentation (non-instance) and another for detecting keypoints for improving instance-level people segmentation.  

% \item We demonstrate how to directly learn pose-conditioned instance segmentation by estimating keypoints and segmentation masks jointly in a deep network.

% \item We introduce an architecture that jointly learns keypoints and segmentation masks, with a \emph{stack} that provides a learned model of how the segmentation can be conditioned on the pose keypoints.

\item{We show 
how the \emph{learned} segmentation can be conditioned on the keypoints by learning 
additional parameters 
% (18 extra parameters) 
specifically for mapping shape to segmentation while training a
DCNN
\emph{jointly} for keypoints and segmentation. }

\item We perform extensive empirical investigation of the proposed Pose2Instance method on the intersection of COCO instance segmentation and COCO keypoints dataset. We show the effectiveness of the pose conditioned deep instance segmentation model by qualitative and quantitative analysis.

\end{itemize}

% \textbf{mdc}: Long-ish contribution list, would any of it work within the main text? Especially for the middle “methods” portion. 

% To the best of our knowledge, we are the first to show that instance-level people segmentation can be achieved through a learnable pose-conditioned segmentation model.

\section{Related Work}
\label{sec:prior-art}
Our work builds upon a rich literature in both semantic segmentation using convolutional neural networks and joint pose-segmentation modeling.\\

\textbf{Semantic and Instance Segmentation}\\
DeepLab \cite{deeplab_chen14semantic} and FCN \cite{fcn_cvpr_15} achieved significant breakthroughs for the challenging task of semantic segmentation using deep convolutional neural networks. 
Subsequently, a set of instance segmentation methods
\cite{rev_inst_seg_cvpr16,mrf_ZhangFU15, itr_inst_seg_LiHM15, propo_free_inst_LiangWSYLY15, Inst_sens_DaiHLRS16, rec_inst_seg_Romera_ParedesT16, ICLR_WS_ParkB15a, Bottom_up_inst_BMVC_2016} 
were proposed, which begin with pixel-wise semantic segmentation and generate instance-level segmentation from them. 
Recently, \cite{TA-FCN_coco_16} achieved the state-of-the-art
performance on the $80$-category instance segmentation using a fully convolutional end-to-end solution. 
Except \cite{itr_inst_seg_LiHM15}, none of these methods look into learning implicit or explicit shapes of different object categories.
% TODO(mdc): This section could use a lot more detail, especially given the absence of page constraints for this version.

\textbf{Human Pose Estimation}\\
Human pose estimation from static images \cite{gram_seg_for_pose_RothrockPZ13, Kohli2008, we_are_family_eccv_EichnerF10, pose_est_review_Liu_2015} or videos \cite{grabcutSensors_2012} with hand-crafted features and explicit modeling gained considerable interest in the last decade.  
Human pose estimation using an articulated grammar model is proposed in  \cite{gram_seg_for_pose_RothrockPZ13}. 
Hern\'andez-Vela \etal \cite{grabcutSensors_2012} proposed Spatio-Temporal GrabCut-based human segmentation that combines tracking and segmentation with hand-crafted initialization. 
% \cite{multiview_shape_pose_eccv_Rhodin16} jointly creates a rigged actor model commonly used for animation – skeleton, volumetric shape, appearance, and optionally a body surface – and estimates the actor’s motion from multi-view video input only.
In \cite{we_are_family_eccv_EichnerF10}, Eichner and 
Ferrari proposed a multi-person pose estimator framework that extends pictorial structures for explicitly modeling interaction between people. A detailed review on pose estimation literature survey is available in \cite{pose_est_review_Liu_2015}.

Recently, convolutional neural networks have been successfully applied for pose estimation from videos \cite{pose_video_LinnaKR16}, human body parts segmentation \cite{part_discovery_gabriel}, and multi-person pose estimation \cite{deepcut_cvpr16,deepercut_eccv16,coco_challenge_16_cmu_pose,coco_challenge_16_GRMI_pose}. Additionally, among the most accurate results are those shown by \emph{chained prediction} \cite{Gkioxar_eccv16}.
% TODO(mdc): Does the below belong in pose estimation? Needs a rewrite, though.
% which estimates pose in static images or videos and using a recurrent LSTM layer \cite{tensorBox16} reported significant improvement on end-to-end human pose estimation.\\

\textbf{Joint Pose Estimation and Segmentation}\\
The most closely related works to this one are those that also seek to jointly estimate human pose and segmentation in static images or videos \cite{Kohli2008,Lim_2013_ICCV,pose_INRIA_Alahari13,Pose_Inria_Seguin15}. 
Kohli \etal \cite{Kohli2008} proposed \emph{PoseCut}, a conditional random field (CRF) framework to tackle segmentation and pose estimation together for a \emph{one} person. The CRF model explicitly combines hand crafted image features and a prior on shape and pose in a Bayesian framework. The prior is represented by the distance transform of a human \emph{skeleton}. 
The \textit{inference} in PoseCut
finds the MAP solution of the energy of the pose-specific CRF. 
The test time prediction finds MAP by doing optimization over different configurations of the latent shape prior.
With 
a \emph{good} initialization 
the inference step requires ~$50$ seconds per frame.
Similar inference strategies for deep models are computationally prohibitive. 

% TODO(mdc): This paragraph is somewhat out of place. Should it go in related work? Should it appear alongside introduction of the distance transform?
% \textbf{Distance Transform for shape-likelihood:}
% Kr{\"{a}}henb{\"{u}}hl and Koltun \cite{gop_eccv_KrahenbuhlK14} also used the signed distance transform of foreground and background seeds for generating object proposal candidates. Recently, \cite{HayderHS16} also used the notion of distance transform independently from us and use it for shape-aware instance segmentation. 
% % They use deconvolutions to create a differentiable version of the distance transform.  
% Unlike \cite{HayderHS16}, we focus more on keypoint-based poses, and the additional training signal that can come from those.

Among other significant efforts towards joint pixel-wise segmentation and pose estimation of multiple people, Alahari and Seguin \textit{et. al.} \cite{pose_INRIA_Alahari13,Pose_Inria_Seguin15} use additional motion and disparity cues from stereo videos. The appearance and disparity cues are generated using HOG features. The pose estimation model \cite{pose_INRIA_Alahari13} is represented as a set of parts, where a part
refers to a patch centered on a body-joint or on an interpolated
point on a line connecting two joints. They learn up to eight mixture components for each part and an articulated pose mask for the mixture components. 

We propose a different and effective framework for incorporating pose prior into deep segmentation models.
The proposed DCNN model consists of additional parameters that are trained/optimized specifically for the mapping of shape to segmentation. 
The Pose2Instance \emph{inference} does not require optimization such as finding the MAP solution. 
The prediction task involves only 
\textit{one} forward pass
through the trained network. 
% Thus the run-time of the inference for Pose2Instance is similar to DeepLab ($<.2$ seconds) 
% which is at least $100 \times$ faster than finding MAP solution as in \cite{Kohli2008}.

\section{Methods}
\label{sec:methods}

Our Pose2Instance approach looks at the problem of incorporating a pose prior into segmentation in two ways. We begin with a constrained environment study where the keypoints are provided by an oracle, and we investigate a way for improving the instance segmentation inference given a state-of-the-art pixel-level person classifier \cite{deeplab_chen14semantic}. Next, we move to a more realistic case where oracle keypoints are not available and propose a framework to train segmentation model directly while benefiting from a pose estimator. 

% We use the notion of distance transform of oracle provided keypoints or estimated keypoint heatmaps as a prior for the people instance segmentation task. 
% We first show how the oracle keypoints can boost existing human semantic segmentation model. Next, we also demonstrate how to directly learn a segmentation model conditioned on keypoints estimation.

% mdc: The following figure can probably already be inferred from Figures 1.1 and 3.4, plus the description of the changes relative to those figures that appear in this section.

% \begin{figure}
% \begin{center} 
%   \includegraphics[width=3.1in]{Pose2Instance_Inference.png}
% \end{center}
%   \caption{Pose2Instance inference in constrained environment for oracle provided keypoints (best viewed in color.}
% \label{fig:Pose2Instance_Inference_with oracle keypoints}
% \end{figure}

\begin{figure*}
\begin{center}
	\includegraphics[width=2.1in]{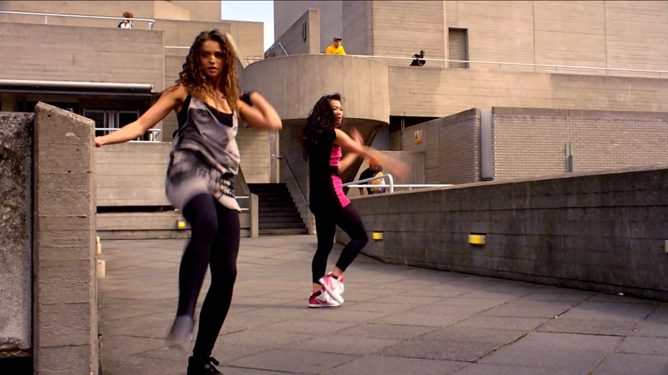} 
    \includegraphics[width=2.1in]{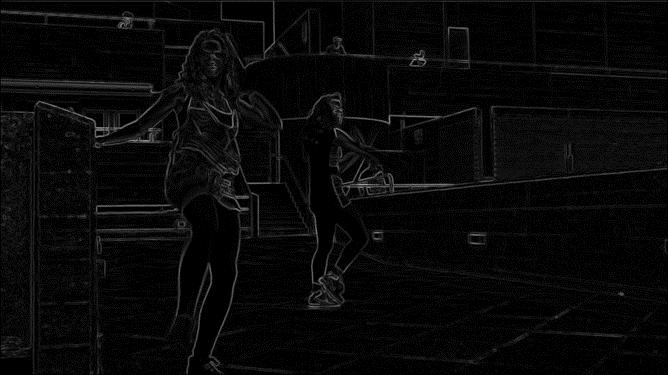}
    \includegraphics[width=2.1in]{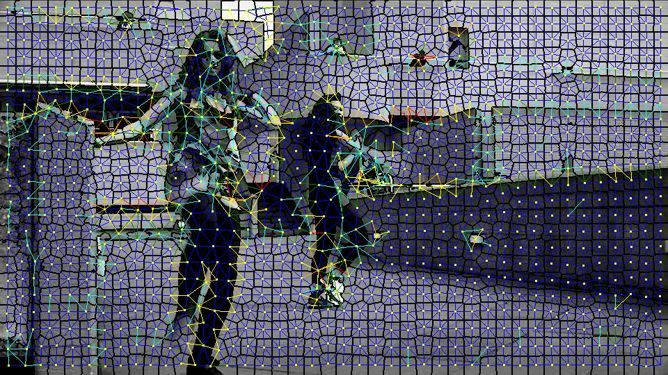}  
\end{center}
\begin{center}	
    \includegraphics[width=2.1in]{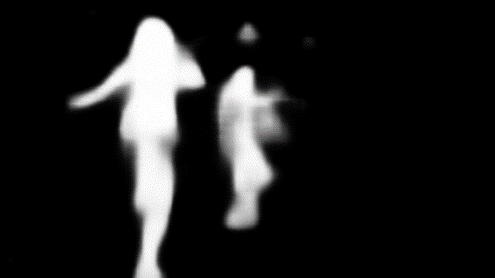}
	\includegraphics[width=0.7in]{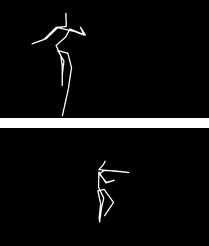}
    \includegraphics[width=0.7in]{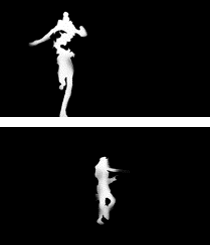}
    % TODO: spacing?
    \includegraphics[width=0.7in]{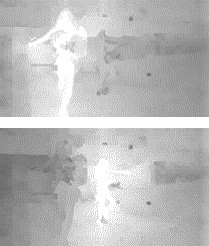}
    \includegraphics[width=2.1in]{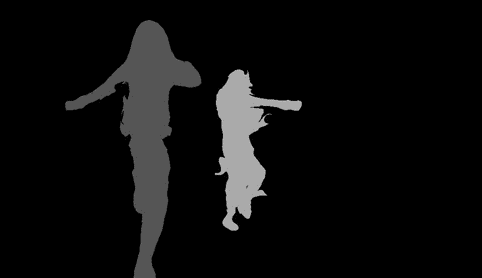}
\end{center}
   \caption{Instance Segmentation with oracle skeleton. Top row: An  image from Inria stereo people segmentation dataset; Sobel edge responses on the image; RAG with edge strength from sobel responses where warmer color means higher weights. Bottom row: DeepLab \emph{person} segmentation; oracle skeleton, masks for distance transform in RAG space, \emph{pose-instance-maps} and the final instance segmentation which is an argmax on point-wise multiplication of pose-instance map and the DeepLab-people score. }
\label{fig:Pose2instance_oracle_stickmen1}
\end{figure*}

% \subsection{Dataset}
% In order to evaluate the pose-conditioned segmentation model, we consider all instances of (non crowd) people from COCO \cite{COCO_eccv14} segmentation dataset where instances also have keypoints ground truth. By comparing the image and instance identifiers from the COCO segmentation and COCO keypoints dataset, we see there exists $45,174$ images in the training dataset and $2,644$ images in the validation dataset. 
% This amounts to $185,316$ and $10,709$ ground truth person instances which have both the segmentation and keypoints annotations in the training and validation split respectively. 
% We call this intersection between COCO instance segmentation dataset and COCO person keypoints dataset as COCO dataset throughout this paper. 

\subsection{Pose2Instance Inference Only}
We first present Pose2Instance within a constrained environment that assumes that the keypoints are provided by an oracle. This allows us to investigate the contribution of the pose prior independent of the other components of the whole system.
In the COCO dataset, $17$ person keypoints along with their corresponding visibility flags are annotated. We will handle these as part of a \emph{skeleton} that links joint keypoints by the corresponding body parts.
% TODO: Should have detailed diagram of the skeleton, for COCO, in the supplement?

In the investigation of the prior alone, with oracle keypoints, we address the inference stage of instance segmentation without any training.
The sole task-specific training is done on the already existing DeepLab \cite{deeplab_chen14semantic} network. 
In the section \ref{oracle_keypoints} below, 
we first fine-tuned this network for person-specific segmentation on COCO, with other labels discarded. This model directly predicts per-pixel probability of the \emph{person} class label for the whole image.
We call this model \emph{DeepLab-people} in this paper.

\subsubsection{Person Instances from Oracle Keypoints}
\label{oracle_keypoints}
We use the notion of a distance transform of the person skeleton \cite{Kohli2008}, generated from the oracle keypoints, as a prior for the instance segmentation task.
For this proof of concept, we follow the below steps.

We create a Region Adjacency Graph (RAG), \mbox{\textbf{G} = (\textbf{V}, \textbf{E})} where the nodes \textbf{V} are superpixels and the weights of the edges \textbf{E} between the nodes depend on the strength of image edges. We obtain superpixels using SLIC \cite{slic12}, and the image edge responses using Sobel operator.
We can define the pose prior as a distribution over the labels of this graph.
Given a superpixel $p \in \textbf{V}$, we can compute a conditional probability it belongs to
a given instance.
For each instance, we color those nodes where the corresponding superpixel contains a part of the human skeleton line that is generated from the oracle keypoints with valid visibility flags. The colored nodes in the RAG represent a foreground binary mask, and are assigned the highest probability of belonging to this instance.
% Each human skeleton from the oracle keypoints selects those superpixels as a part of its corresponding binary mask where a superpixel contains a part of the skeleton line.
For each such binary mask corresponding to each person, we apply distance transform in the RAG using Floyd-Warshall \cite{Floyd-Warshall} shortest paths algorithm. A point-wise softmax of this distance transform then represents the likelihood of each person's gross shape. We call this shape-likelihood the \emph{pose-instance map}. For an image of height $h$ and width $w$, with $n$ oracle instances, the shape of pose-instance map is $h\times  w \times n$.  
Figure \ref{fig:Pose2instance_oracle_stickmen1} shows these intermediate steps of generating the RAG, its nodes and the weights of its edges, the oracle skeletons and the instance segmentations. \\

\textbf{Instance-level to Image-level inference}:
Element-wise multiplication of this pose-instance map and the DeepLab-people score  generates instance heatmap of size $h \times w \times n$. Here, $h$, $w$ and $n$ denote the height and width of the image, and the oracle-provided number of person instances respectively. An argmax on instance heatmap produces the final instance segmentation on the image.
%Figure \ref{fig:Pose2Instance_Inference_with oracle keypoints} shows the Pose2Instance inference stage with oracle provided keypoints. The inference step takes the whole image and all oracle provided keypoints as inputs and produces the person instance segmentation on the whole image.

\begin{figure}
\begin{center}
	\includegraphics[width=1.6in]{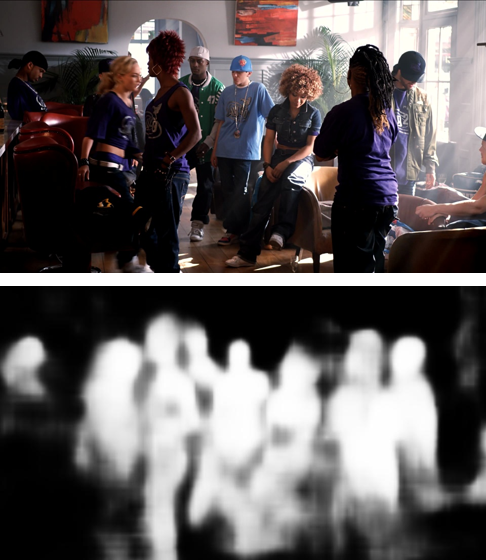} 
    \includegraphics[width=1.6in]{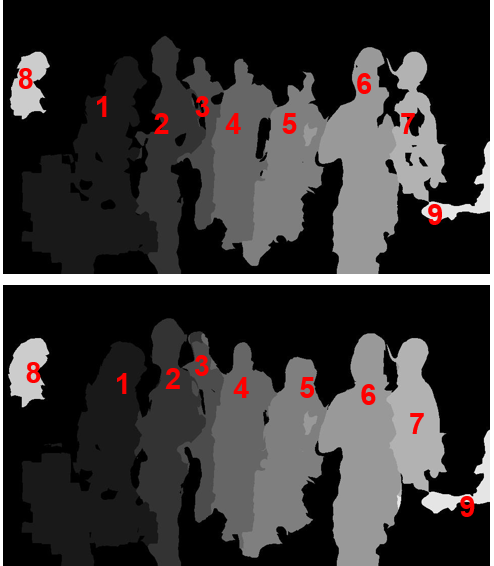}  
\end{center}
   \caption{Pose2Instance inference with oracle skeleton. Top row: An image from Inria stereo person segmentation containing 9 persons; Instance classification by argmax on pose-instance map;
Bottom row: DeepLab-people score, and the \emph{Pose2Instance} inference output which is an argmax on \emph{instance heatmap} generated by fusing DeepLab-people and \emph{pose-instance map}.}
\label{fig:Pose2instance_oracle_stickmen2}
\end{figure}

Figure \ref{fig:Pose2instance_oracle_stickmen2} shows intermediate results for the inference step in this constrained setup. There are $9$ persons in this image. Combining DeepLab-people score with pose-instance map improves the instance segmentation quality over the pose-instance map.  
Quantitative results in section \ref{sec:inference_with_oracle} show that person keypoints represented as the distance transform can be an excellent source of additional domain knowledge for improving people instance segmentation.  

% * <subarna.tripathi@gmail.com> 2016-10-26T05:16:20.660Z:
%
% I have visual results of distance transform with fast sweeping methods.  But don't have the objective result for the entire COCO validation dataset. 
%
% ^.

% \begin{figure*}
% \begin{center}
% 	\includegraphics[width=2.1in]{inria2_part1.png} 
% %     \includegraphics[width=2.1in]{oracle_inria2_stickmen.png}
%     \includegraphics[width=2.1in]{inria2_part2.png}  
% \end{center}
%    \caption{Instance Segmentation from Oracle Skeleton. From top to bottom and left to right. A frame from Inria stereo people segmentation dataset with 9 people in close proximity; Instance classification by likelihood from oracle skeletons; Combining DeepLab human segmentation output and Pose2Instance result.}
% \label{fig:Pose2instance_oracle_stickmen2}
% \end{figure*}

\subsubsection{Person Instances from Oracle Bounding Boxes}
As a baseline, we take the approach of snapping the pixel-level DeepLab-people score at oracle bounding boxes for the 
% $2,644$ 
COCO validation images.
Though this bounding box approach does not comply with the relative depth ordering or visibility of one instance over another, the method still can be used as a reasonable baseline to compare the performance of Pose2Instance inference. \\
% We use the same instance to full-image inference methodology to generate full-image results.

We performed similar experiments with fast-sweeping \cite{fast_sweep_Weber_2008} based distance transform on pixel grid for reducing complexity using single-pixel width skeleton as the binary mask. However, the distance transform produces worse result comparing with the specified non-grid RAG approach.

\subsection{Learning Pose2Instance} \label{shape likelihood}

After this proof of concept in the inference stage in a controlled setup with oracle keypoints, we move to a more realistic scenario where ground truth keypoints annotations are unavailable and we strive for learning a segmentation model by jointly optimizing for the segmentation and pose. 

Our proposed network has a DeepLab-style architecture \cite{deeplab_chen14semantic}. This is a modified VGG network \cite{vgg_SimonyanZ14a} that uses atrous convolution with hole filling \cite{deeplab_chen14semantic} and replaces fully-connected layers by fully-convolutional layers. The baseline model is a $2$-class \emph{DeepLab-people} model.
To construct this model, we start with the publicly available Deeplab model trained on the PASCAL VOC dataset, and fine-tune it for predicting only people on the COCO training instances. 
The second and third exploratory architectures involve two output layers each, a segmentation output and a $17$-channel heatmap for pose estimation output. 
The first among them \emph{Pose and Seg} is a multitask model, where the two parallel output layers share the parameters up to previous convolutional layers. The $2$-class segmentation layer and the $17$-class pose estimation output layers use cross-entropy loss after softmax and sigmoid activations respectively. 
%Note: I also tried l2-loss for the pose heatmaps, but both multitask and cascaded models performed worse than the COCO finetuned DeepLab. There could be issue with balancing classification and regression loss weights.
The later one, \emph{Pose2Seg}, is 
a cascaded model, where the $17$-channel keypoints heatmap is followed by an $1\times1$ convolution to generate the shape likelihood. Segmentation feature maps from the last layer is combined with the above shape likelihood, and the softmax segmentation is trained.   
Comparing with the segmentation only model, the cascaded model has only $18$ extra parameters for learning the $1\times1$ convolutional kernels. 
$17$ parameters are for the key-points heatmaps, and $1$ for shape likelihood.  

\begin{figure*}
\begin{center}
	\fbox{\includegraphics[width=3.2in]{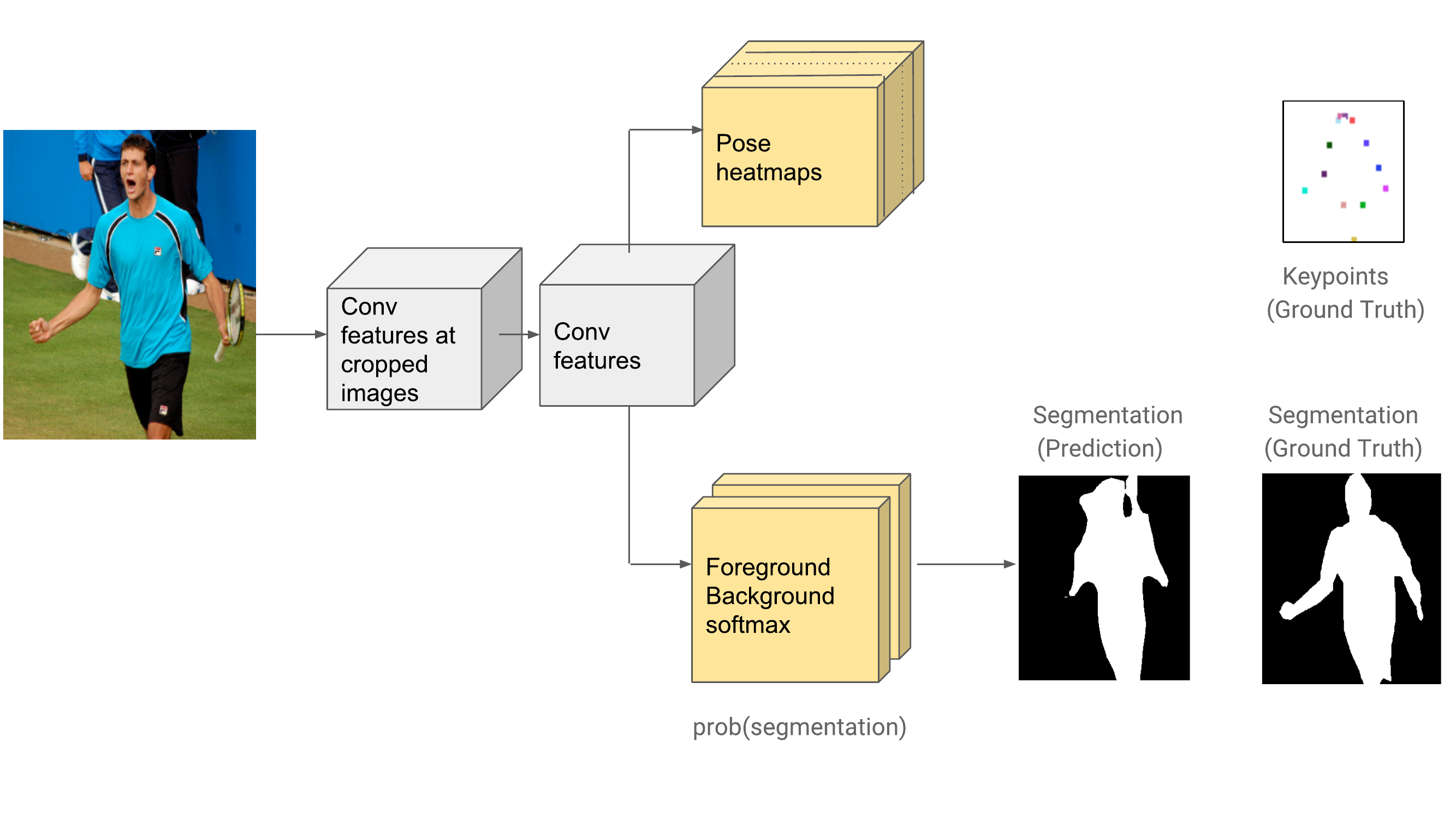}}   %{segonly_3}}
    \fbox{\includegraphics[width=3.2in]{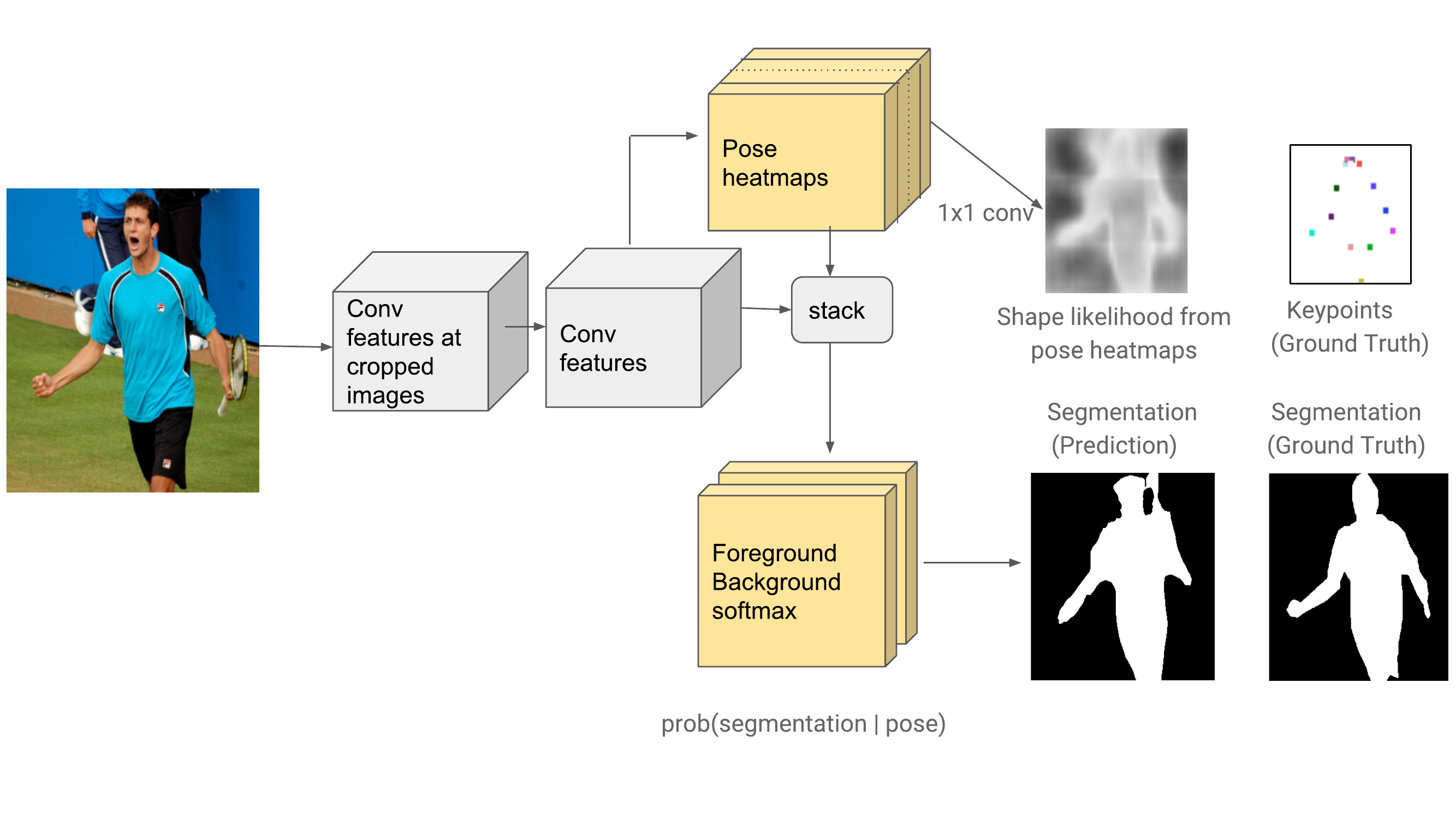}} 
\end{center}
   \caption{Pose2Instance: architecture variation for joint pose and segmentation learning. Left: Multitask model where pose estimation and segmentation are two parallel output paths. Right: Cascaded Model where pose dedicated parameters are \emph{learned} for mapping pose to segmentation.}
\label{fig:Learning Pose2Instance}
\end{figure*}

Figure \ref{fig:Learning Pose2Instance} shows the two above mentioned architectures. The \emph{stack} operation is a $1\times 1$ convolution on the estimated $17$-channel pose heatmap. Its output can be used as the gross shape-likelihood of a person based on the estimated keypoints. 
% Thus, in the cascaded model we can say that the model learns to predict the instance segmentation probability conditioned on estimated distance transform, thus conditioned on the estimated keypoints.
In the cascaded model, the segmentation output is directly conditioned on the pose heatmap. As we also see in Fig \ref{fig:Pose2instance_learn results2}, the $1\times 1$ convolution on the pose heatmap preserves the general notion of shape of a person from its keypoints, the segmentation model thus  
can be thought of conditioned on the latent shape of a person. 

In the Pose2Instance framework, we try to improve the segmentation accuracy from both keypoints and segmentation supervision. In particular, a model learned with one supervisory signal \emph{pose-estimation} acts as a prior to the model learned with another supervisory signal \emph{segmentation}.  
Different sources of supervision have proven to be useful for learning segmentation. 
% either single output or multitask model.
For example, ScribbleSup \cite{scribble_sup_LinDJHS16} performs semantic segmentation from additional scribble based supervision broadly in grabcut \cite{grabcut_MSR} like framework. In \cite{Bearman16}, Bearman \etal discussed various levels of supervisions such as pixel-level strong supervision, and sparse point-level supervision for semantic segmentation. 
Our method is substantially different from these since none of the above specifically addresses (instance) segmentation problem with one as a prior to the other.

\section{Results}
\label{sec:results}

We implement the Pose2Instance model using TensorFlow-Slim. We train the model on specified COCO training instances. We initialize the model from DeepLab-people and continue training for $20,0000$ iterations using stochastic gradient descent with mini-batch size of $16$ and momentum $0.9$.

\subsection{Pose2Instance in a Constrained Setup} \label{sec:inference_with_oracle}
\begin{figure*}
\begin{center}
	\includegraphics[width=6in]{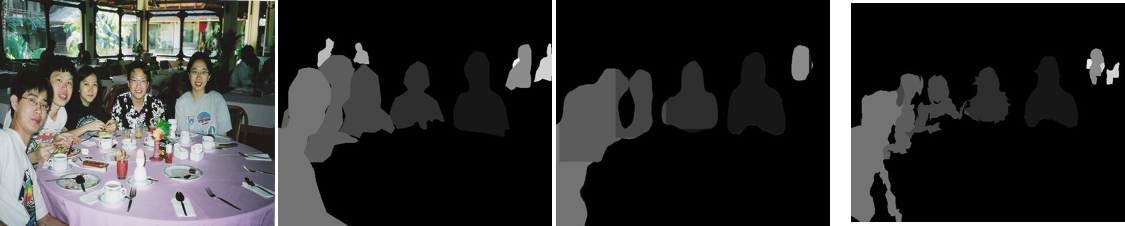} 
\end{center}
\begin{center}
	\includegraphics[width=6in]{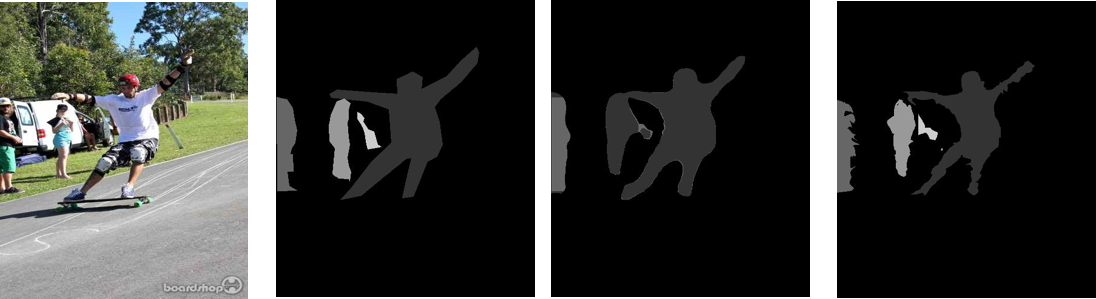} 
\end{center}
   \caption{Pose2Instance with oracle bounding boxes vs oracle keypoints. From left to right: A frame from COCO keypoints dataset; Ground truth instance segmentations; Baseline instance segmentation from oracle bounding boxes and Pose2Instance inference from oracle keypoints.}
\label{fig:Pose2instance_oracle_stickmen_eval}
\end{figure*}

In order to analyze the Pose2Instance inference with oracle keypoints, we use COCO validation images.
While table \ref{table:oracle_eval_baseline} shows the performance of Pose2Instance inference with oracle keypoints,
figure \ref{fig:Pose2instance_oracle_stickmen_eval} shows some qualitative results 
% of people instance segmentation on COCO validation dataset 
comparing with the oracle bounding box baselines. The figures show that for overlapping person instances, the proposed pose prior significantly outperforms a baseline using the bounding box as an ad-hoc prior.  \\

\begin{figure}
\begin{center}
	\includegraphics[width=3.1in]{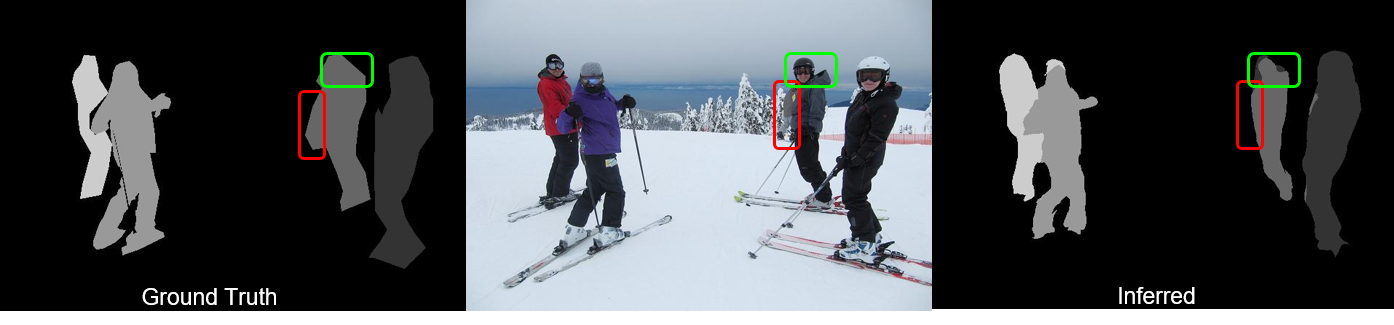} 
\end{center}
\begin{center}
	\includegraphics[width=3.12in]{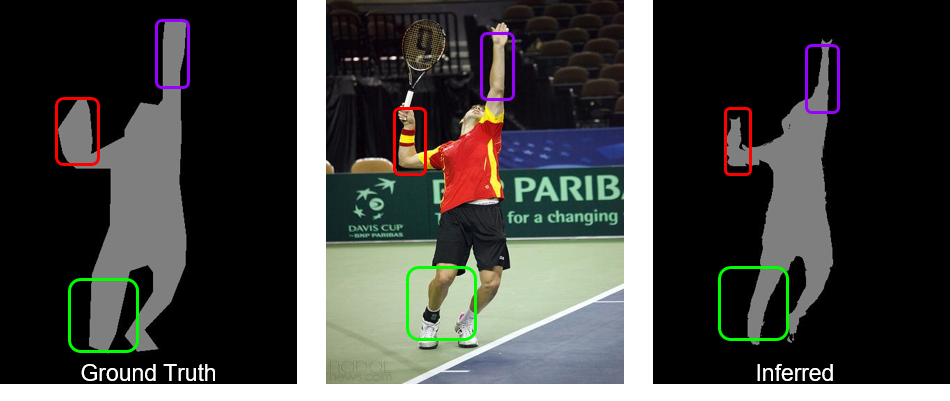} 
\end{center}
   \caption{
%    Pose2Instance with oracle keypoints. 
From left to right: Ground truth instance segmentations; Corresponding image from COCO keypoints dataset; Pose2Instance inference with oracle keypoints. 
%    Top row: Two different persons were incorrectly annotated as a single instance (marked in red). 
%    Bottom row: segmentation annotation is far from being perfect (areas marked in different colors). Keypoints supervision helps in improving segmentation prediction. 
% Human keypoints ground truth is easier to collect than precise segmentation masks.
Colored boxes show errors in segmentation ground truth that are corrected using our keypoint conditioned model.
}
\label{fig:coco_gt_analysis}
\end{figure}

\begin{table}[h] \label{table:oracle_eval_baseline} 
\centering
%\scriptsize
\begin{tabular}{lll}
%\toprule
\multicolumn{3}{c}{\textbf{Instance Segmentations on COCO validation Images}} \\ 
\toprule
    Methods & $AP^r 0.5$ & $AP^r$ \\
            & \@ IoU=0.5   &   \@ IoU=\lbrack0.5 to 0.9\rbrack \\
    \midrule
    DeepLab+Oracle BB & 0.437 & 0.252\\ 
    DeepLab+Oracle keypoints & $\mathbf{0.533}$ & $\mathbf{0.283}$\\
    \midrule
    FAIRCNN\cite{FAIRCNN_ZagoruykoLLPGCD16} & 0.504 & 0.206 \\
    CUHK\cite{coco_leaderboard_1026} & 0.478 & 0.214 \\
\bottomrule
\end{tabular}
\caption{
% Pose2Instance inference on the COCO validation dataset. 
% DeepLab-people model is combined with pose-instance map generated either with oracle bounding box or oracle keypoints. 
Oracle keypoints provides $10\%$ to $12\%$ relative improvement over oracle bounding box case at various IOU thresholds when applied on DeepLab-people segmentation model. Results are shown from COCO Leaderboard for FAIRCNN\cite{FAIRCNN_ZagoruykoLLPGCD16} and CUHK\cite{coco_leaderboard_1026} that also use VGG as the base network. }
\end{table}

The inference stage which consists of combining the existing semantic segmentation model and the oracle keypoints outperforms the oracle bounding box case by $10\%$ relative improvement. 
FAIRCNN\cite{FAIRCNN_ZagoruykoLLPGCD16} and CUHK\cite{coco_leaderboard_1026} are the instance segmentation models that also use VGG as the base network. We include their instance segmentation results only on \emph{'person'} category from COCO detection challenge Leaderboard as references. 
Newer models from the Leaderboard use more powerful ResNet in their backend, so are not directly comparable.

\begin{figure*}
\begin{center}
	\includegraphics[width=1.24in]{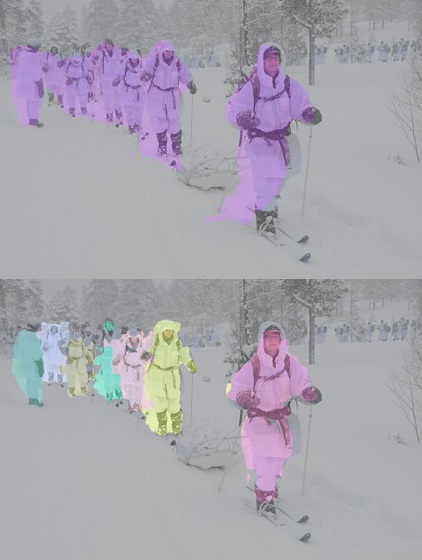} 
    \includegraphics[width=1.1in]{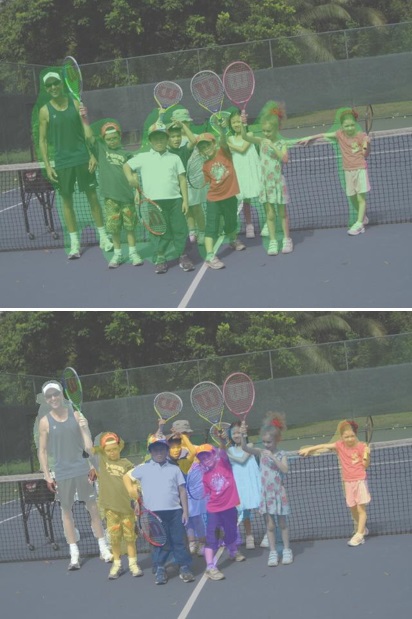} 
    \includegraphics[width=1.47in]{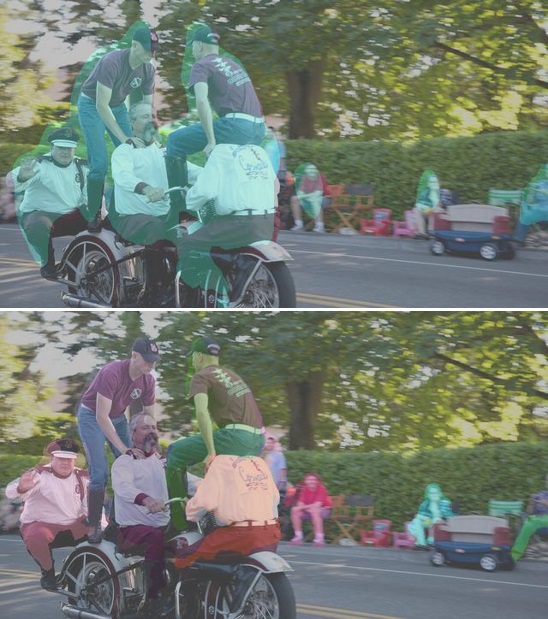} 
    \includegraphics[width=1.15in]{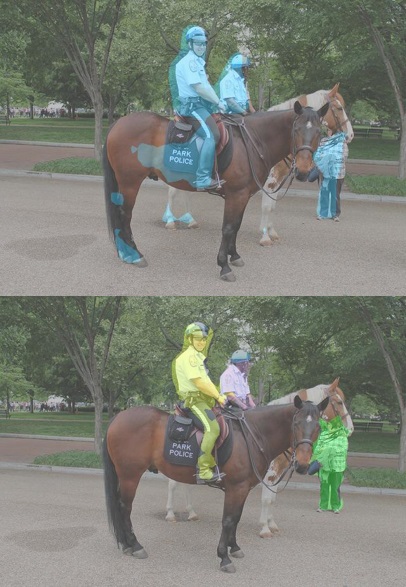}
    \includegraphics[width=1.11in]{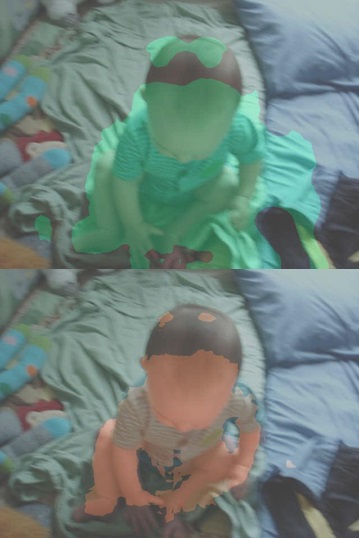} 
\end{center}
   \caption{Pose2Instance in a constrained setup. Top: DeepLab person segmentation. Bottom:Pose2Instance inference from oracle keypoints on COCO evaluation dataset. (best viewed in color)}
\label{fig:Pose2instance_oracle_qualitative results}
\end{figure*}

Figure \ref{fig:Pose2instance_oracle_qualitative results} shows qualitative results of instance segmentation on COCO validation dataset in such constrained environments. We note that this method only applies during the inference step with oracle keypoints and does not involve any training. 
Human keypoints ground truth is easier to collect than precise segmentation ground truth. Figure \ref{fig:coco_gt_analysis} shows how the errors in the segmentation ground truth can be corrected with our pose-conditioned segmentation model. 

\subsection{Pose2Instance in Realistic Environments}

After validating the effectiveness of the inference with keypoint-specific distance transform, we evaluate the proposed Pose2Instance model on COCO validation instances in a more realistic environment where oracle keypoints are unavailable. We assume the availability of oracle bounding boxes. The model estimates the keypoints and segmentations at all instances.

\begin{figure*}
\begin{center}
	\includegraphics[width=5.5in]{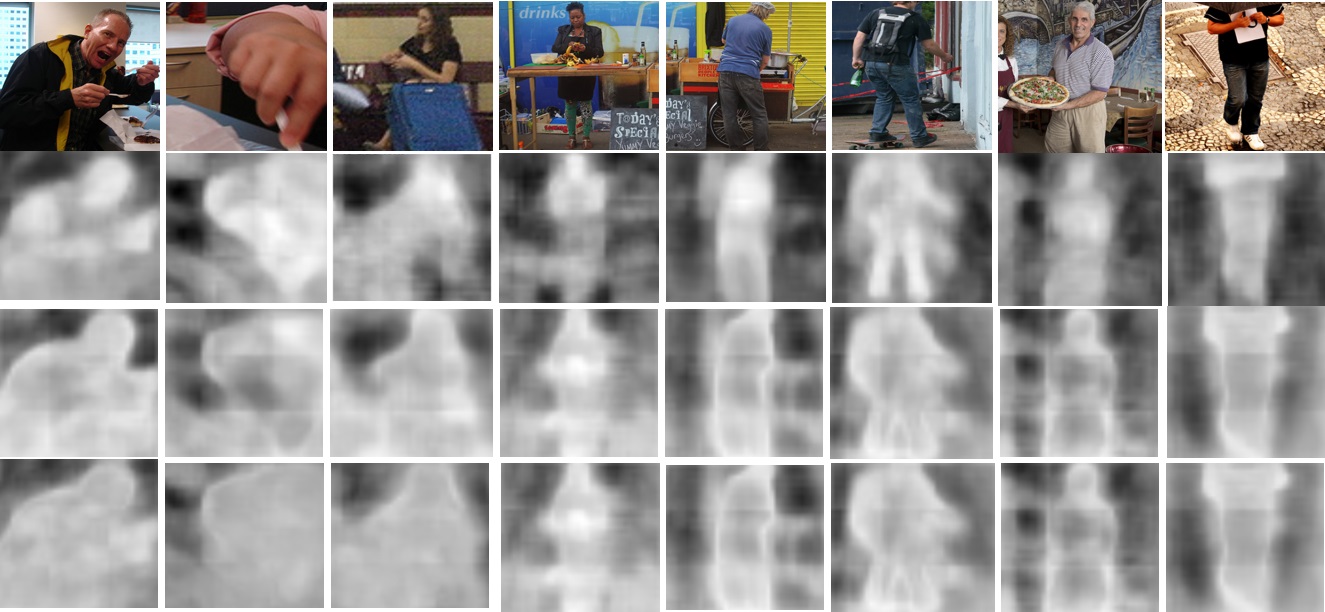} 
\end{center}
   \caption{Qualitative results for shape likelihood from pose estimation. Top to Bottom: Instances from COCO validation dataset; visualization of intermediate latent shape likelihood for (i) \emph{Pose Only model}; (ii) \emph{Multitask model} and (iii) \emph{Cascaded model} respectively. \emph{Pose Only} model produces high likelihood around the keypoints; whereas other two joint models learns to capture the overall person contour shape.}
\label{fig:distance_transform_qualitative results}
\end{figure*}

\begin{table}[h] \label{table:pose2instance_learn} 
\centering
%\scriptsize
\begin{tabular}{lll}
%\toprule
\multicolumn{3}{c}{\textbf{Instance Segmentations without Oracle Keypoints}} \\ 
\midrule
    Methods & $AP^r $ & $AP^r$ \\
            & IoU=0.5   &   IoU=\lbrack0.5 to 0.95\rbrack \\
    \midrule
%     Baseline DeepLab & 0.36 & 0.17\\ 
    DeepLab Seg only & 0.79 & 0.38\\ 
%     \midrule
	Multitask: Pose and Seg & 0.80 & 0.40\\ 
    Cascaded:  Pose2Seg & $\mathbf{0.82}$ & $\mathbf{0.42}$\\   
\bottomrule
\end{tabular}
\caption{ Evaluation of segmentation accuracy on instances from COCO validation dataset. 
Joint pose estimation and segmentation outperforms the \emph{segmentation only} model. Pose2Instance \emph{cascaded} model achieves improved accuracy over the \emph{multitask} model. Overall, relative improvement from \emph{segmentation only} model is $3.8\%$ to $10.5\%$ at various IOU thresholds.}
\end{table}

In Table \ref{table:pose2instance_learn}, we show the comparative segmentation performance evaluation for the proposed Pose2Instance method without oracle keypoints. Average Precision at $0.5$ IOU improves by $3\%$ over the \emph{segmentation only} model and $2\%$ over the multitask model. 
The corresponding improvements for the $[0.5, 0.9]$ IOU are $4\%$ and $2\%$ respectively.

In terms of relative improvements, at $0.5$ IoU and $[0.5, 0.9]$ IoU, 
the pose-conditioned segmentation model improves the $AP^r$ by $3.8\%$ and $10.5\%$ over the \emph{segmentation only} model [Table \ref{table:pose2instance_learn}] respectively. 
% Given the fact that all models are trained on the same dataset, and same 
This demonstrates the proof-of-concept of how to incorporate pose prior effectively into deep segmentation model. 

\begin{figure*}
\begin{center}
	\includegraphics[width=6.5in]{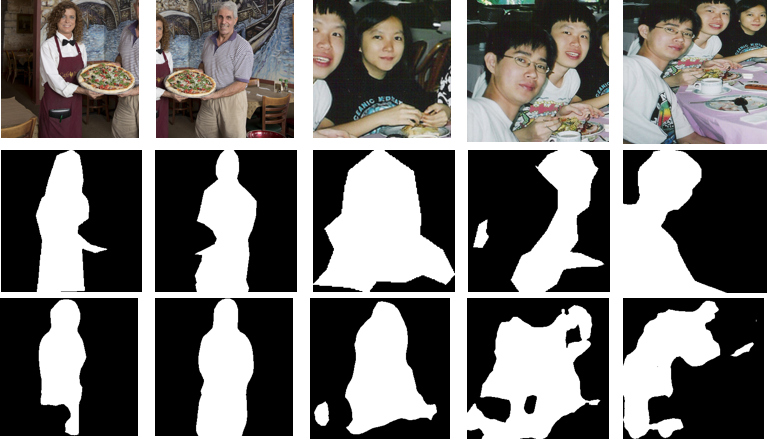} 
\end{center}
   \caption{Pose2Instance without oracle keypoints. Top row: Instance bounding boxes of COCO validation images. Middle row: Ground truth segmentation at instance level. Bottom row: predicted segmentation masks for the instance bounding boxes. Bounding boxes contain multiple full or partial person instances. While the first three columns show successful instance segmentation results, the last two examples show \emph{yet to improve} segmentation results due to failure of the pose-estimator output in the VGG based Pose2Instance model. 
Improving the pose-estimator can improve the accuracy of the pose-conditioned segmentation model.   
% results on challenging examples.
}
\label{fig:Pose2instance_learn results2}
\end{figure*}

Figure \ref{fig:Pose2instance_learn results2} shows qualitative results on some challenging examples. These rectangular regions contain one or more partial person instances in addition to the primary person instance. We see that the Pose2Instance model learns to produce instance segmentation only for the intended one. The last two figures are examples of most difficult cases with many people in close proximity, and the Pose2Instance predictions are far from being ideal due to the current limitation of the VGG-based pose-estimator output.

% still it thrives for segmenting only the desired instance region even with multiple persons in close proximity.
In this work, we assess the effectiveness of pose conditioned segmentation performance, and did not evaluate the parallel key-points estimation output. We performed some qualitative analysis for the pose estimation output from the described multitask and cascaded models. Additionally, we implemented another vanilla pose estimator model with the same network except the segmentation output. We call this model a \emph{PoseOnly} model which is optimized only for $17$-class pose-estimation problem. Our subjective analysis of the latent shape likelihood of person include \emph{PoseOnly}, \emph{multitask} and \emph{cascaded} models. 
Figure \ref{fig:distance_transform_qualitative results} shows some visualizations of the latent shape likelihood (section \ref{shape likelihood}) on some COCO validation images. 
% \squeezeup

% \subsection{Pose2Instance Full System}
\section{Discussions}

%future work1 : evaluate on whole image instead of oracle ground truth boxes ?
%future work2 : use other improved pose-estimation methods in future ?

% We first show how the oracle keypoints can boost existing human semantic segmentation model. Next, we also demonstrate how to directly learn a segmentation model conditioned on keypoints estimation in an end-to-end-framework. The analysis also validates the idea of combining even two existing different models one for people segmentation and another for detecting keypoints for towards improving instance-level segmentation. 

Our experiments suggest that human pose is a useful domain knowledge even atop state-of-the-art deep person segmentation models. 
We show that in a constrained environment with oracle keypoints, at various  IOU thresholds,
the instance segmentation accuracy achieves $10\%$ to $12\%$ relative improvement
over a strong baseline with oracle bounding boxes without any training.
In a more realistic environment, without the oracle keypoints, the proposed Pose2Instance deep model achieves relatively $3.8\%$ to $10.5\%$ higher segmentation accuracy than the strongest baseline of a
deep network trained only for segmentation.
% The success of oracle keypoints or estimated keypoints heatmaps for segmentation suggest that, improving keypoints estimation may lead to even  better instance segmentation.  

Our proposed method 
% extends an existing segmentation
% architecture, 
% and 
is applicable to \emph{any} such architecture that shares
the necessary properties of the Deeplab model.
Models optimized for the segmentation task, including the one covered in our
experiments and future better-performing segmentation models, could
potentially incorporate the same methodology to utilize pose information.

While at present
we show results on images, likely similar dynamics are embedded in videos.
Human keypoints ground truth is easier to collect than precise segmentation masks. Thus, a pose conditioned segmentation model can be more powerful for person instance segmentation for natural scenes where people tend to appear in groups, have dynamic interactions, and partial occlusions. 
This work represents a first step towards embedding pose into segmentation in complex scenes. An exploratory follow-up work can include investigation on incorporating keypoints based dynamic person model into video segmentation. 

% Human keypoints are the most natural representation of a dynamic person model in video. For our futuristic goal of people instance segmentation from such dynamic person models, we want to explore its fundamental stepping stone \ie image-based segmentation conditioned on human pose. 

% \subsection{Acknowledgments}
\section*{Acknowledgments}

% TODO: any funding acknowledgments by Subarna and Serge?

The authors would like to thank George Papandreou and Tyler Zhu for their extensive assistance and insightful comments.

% %%%%%%%%%%%%%%%%%%%%%%%%%%%%%%%%%%%%%%%%%%%%%%%%%%%%%%%%%%%%%%%

{\small
\bibliographystyle{ieee}
\bibliography{egbib}
}

\end{document}